\documentclass[lettersize,journal]{IEEEtran}
\usepackage{amsmath,amsfonts}
\usepackage{algorithmic}
\usepackage{algorithm}
\usepackage{array}
\usepackage[caption=false,font=normalsize,labelfont=sf,textfont=sf]{subfig}
\usepackage{textcomp}
\usepackage{amsmath}
\usepackage{stfloats}
\usepackage{url}
\usepackage{verbatim}
\usepackage{graphicx}
\usepackage{cite}
\usepackage{amsmath}
\usepackage{booktabs} 
\usepackage{tabularx}
\usepackage{tabulary}
\usepackage{CJKutf8}

\begin{document}

\begin{CJK}{UTF8}{gbsn}

\title{An Enhanced Low-Resolution Image Recognition Method for Traffic Environments}
\author{~\IEEEmembership{Zongcai Tan, Zhenhai Gao}
\thanks{The authors are with the college of Automotive Engineering,Jilin University, Changchun, China (e-mail: tanzc@jlu.edu.cn, gaozh@jlu.edu.cn).}}

\maketitle

\begin{abstract}
Currently, low-resolution image recognition is confronted with a significant challenge in the field of intelligent traffic perception. Compared to high-resolution images, low-resolution images suffer from small size, low quality, and lack of detail, leading to a notable decrease in the accuracy of traditional neural network recognition algorithms. The key to low-resolution image recognition lies in effective feature extraction. Therefore, this paper delves into the fundamental dimensions of residual modules and their impact on feature extraction and computational efficiency. Based on experiments, we introduce a dual-branch residual network structure that leverages the basic architecture of residual networks and a common feature subspace algorithm. Additionally, it incorporates the utilization of intermediate-level features to enhance the accuracy of low-resolution image recognition. Furthermore, we employ knowledge distillation to reduce network parameters and computational overhead. Experimental results validate the effectiveness of this algorithm for low-resolution image recognition in traffic environments.

\end{abstract}


\section{Introduction}
\IEEEPARstart{I}{n} recent years, the rapid development of computer technology, electronic technology, and automatic control technology has directed modern transportation tools towards intelligent vehicles. The intelligent driving system, with information technology as the core, has four main components: environmental perception, precise positioning, path planning, and execution by wire. The environmental perception system holds a crucial position in the exchange of information between intelligent driving vehicles and their surroundings. It is the foremost and most challenging part of achieving intelligent driving. Environmental sensing technology aims to gather data using sensors that detect the location of roads, vehicles, pedestrians, and obstacles. This information is then sent to the control system, allowing intelligent driving vehicles to replicate human perception. By doing so, they become more aware of their own driving context and surroundings which enables them to make better decisions \cite{1}. Machine vision sensor-based vehicle image recognition is a vital component of the environmental perception system, serving as the foundation for environmental information gathering, driving alerts, and recognition of ADAS functionalities.\par
Due to the tremendous achievements of deep learning in computer vision and image processing, image processing based on deep learning has emerged as a crucial theoretical foundation for intelligent machine vision in driving. By virtue of the deep network's powerful feature extraction capabilities, it can comprehensively extract high-level semantic information from the image. Jiang et al. \cite{2} utilized features extracted from a deep convolutional network trained on the vehicle type database to enable accurate vehicle detection and extraction from images, which can be further employed for recognition tasks. By utilizing the technique of transfer learning, Yang et al. \cite{3} achieved positive outcomes in large vehicle data sets through recognizing deep features from the Imagenet training model via fine-tuning and other methods. These methods demand superior image quality, and commonly necessitate high-resolution images captured from the standard vehicle perspective. However, collected images in real traffic environments are frequently affected by occlusion interference at intersections, turning intersections or traffic congestion scenes which may involve a mixture of both people and vehicles. Furthermore, captured images are typically characterized by low definition owing to factors such as light, weather, and other external influences, which may also result in inaccurate focusing and increased noise levels. Additionally, hardware costs, compatibility issues, and other practical application problems result in low pixel counts for vehicle-mounted cameras. As a result, the collected images are in low resolution, and obtaining and learning effective feature information from them is challenging. This increases the demands on the robustness and adaptability of vehicle image recognition methods.\par
Currently, a variety of well-developed methods exist for low resolution image recognition. For instance, the super-resolution image reconstruction approach \cite{4}\cite{5} transforms low resolution images into high-resolution images before conducting recognition or classification tasks. Deep learning-based super-resolution image reconstruction tasks typically operate in two modes. During the training process, the feature mapping of low-resolution image and high-resolution image are respectively input into the model, and the relationship between them is connected by using the deep network feature learning ability, realizing the unified mode of end-to-end model.  For instance, Chao and colleagues \cite{6} utilised a deep learning model to acquire the mapping correlation between low-resolution and high-resolution images to accomplish super-resolution image reconstruction, which yielded the highest performance during the same period. Wang et al. \cite{7} utilised both low-resolution and high-resolution images as input for their model and learned the spatial mapping relationship between them through a deep network. This technique effectively solves the issue of low-resolution image recognition and has great potential in the recognition of low-resolution faces and characters.  Another approach involves providing feedback to the prediction relationship between the low-resolution image and the real high-resolution image, guiding and supervising the entire model. This method utilises only the low-resolution image as input and adjusts the real model through interaction with the high-resolution image and the real image. An example of this approach is the super-resolution task based on the generative adversarial network \cite{8}. \par
The challenge of the super-resolution reconstruction approach is to establish a sensible assumption about the high-resolution image's degradation process into a low-resolution image. Subsequently, its intrinsic characteristics and relationships must be identified based on this assumption. This technique delivers high accuracy for specific applications like video surveillance, character recognition and related domains, nonetheless, such occasions require substantial prior knowledge. Higher-resolution images can be obtained from a single image through various methods, including interpolation and other techniques for enhancing texture areas. However, these methods only change the visual effect and do not provide additional feature information, which may not be of significant value for effective image feature extraction \cite{9}. The subspace method for recognising low-resolution images is distinguished by its ability to extract discriminative features from such images, and map them to the same subspace as high-resolution images' features, for optimal classification and recognition outcomes that lead to significantly improved rates of recognition.\par
Owing to the limited capacity of the vehicle-mounted machine vision perception module to capture only low-resolution images in complicated real-world surroundings, our paper analyzes the limitations of this function in the external environment, algorithm structure and hardware performance, and proposes an algorithm model based on deep learning neural network. Based on the characteristics of low-resolution images, we have developed a low-resolution image recognition network that utilizes the residual network structure. This network incorporates a double-branch structure and the common feature subspace technique to precisely identify low-resolution images. Furthermore, we have formulated a low-resolution image recognition algorithm employing the double-branch residual network. The residual module, double branch network and common feature subspace method are utilised to enhance the network structure. Compared with the present network model, the resultant accuracy and model complexity comprehensively of the new network structure with optimised parameters are improved, ultimately catering to the requirements of intelligent vehicle applications.

\section{Neural Networks for \\ Low Resolution Image Recognition}
 One of the key challenges in addressing low-resolution image recognition lies in the extraction of discriminative intrinsic features from images. However, low-resolution images pose significant difficulties for feature extraction due to a reduction in effective features and the presence of degradation and noise interference compared to high-resolution counterparts. After conducting comparative experiments involving various network architectures such as VGG, plainnet, and ResNet, it was evident that residual networks (ResNet) stood out for their exceptional characteristics, including enhanced information propagation, the promotion of feature reuse, and integration. Consequently, this paper opted for residual modules to extract features from low-resolution images and systematically investigated their specific impacts.\par
Residual networks, originally proposed by Kaiming He et al. \cite{10}, tackle the challenges of network degradation and gradient vanishing in deep networks through skip connections between layers, enabling the construction of extremely deep network architectures.\par

\begin{figure}[h]
\centering
\includegraphics[width=3in]{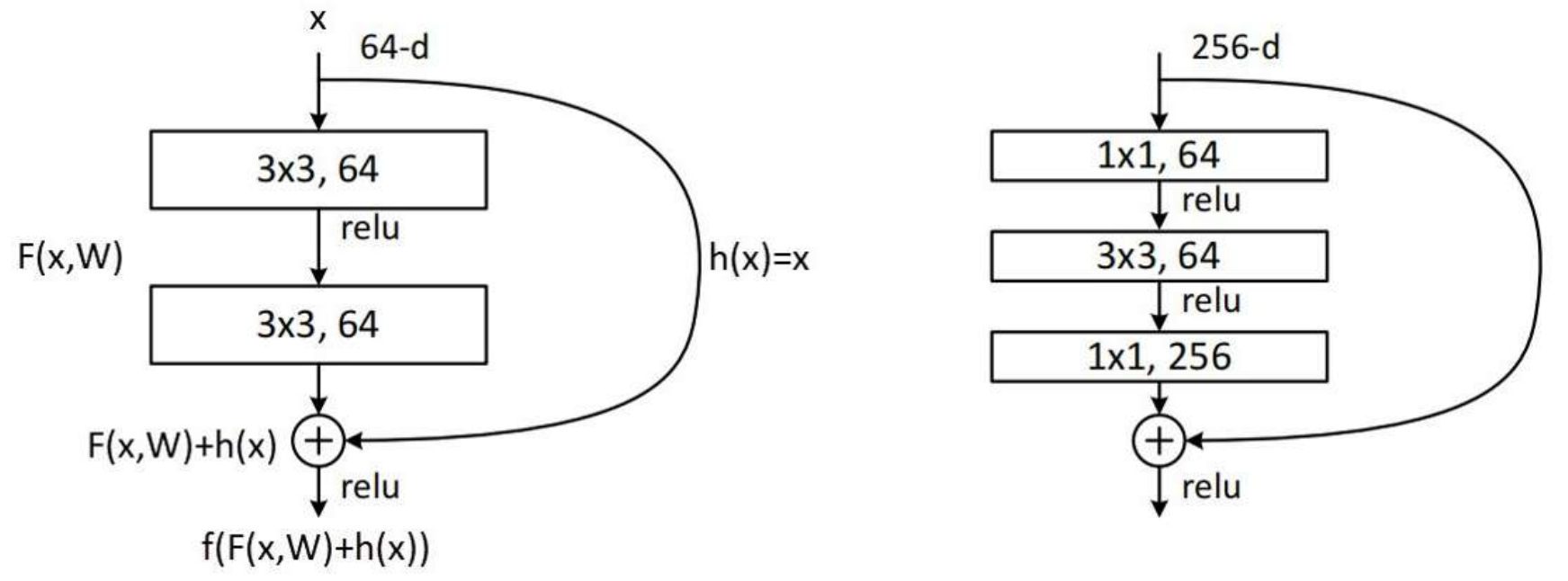} %
\caption{Two fundamental residual module structures \cite{10}.}
\label{fig_1}
\end{figure}

The structure of a single residual module is illustrated in fig.~\ref{fig_1}. A residual module comprises skip connections and residual mappings. Skip connections refer to input features bypassing non-linear convolutional layers and being directly output. Residual mappings denote the results obtained from conventional non-linear convolutional operations on input features. Let $x^l$ represent the input features for the $l$-th residual module, $W^l$ be the corresponding weight matrix (subscript $l$ is omitted for brevity), $h$ denotes the skip connection mapping function (here, it is an identity mapping, hence $h\left(x\right)=x$), and $f$ represents the activation function applied after feature addition (here, ReLU is used). Assuming the desired network layer mapping is denoted as $H\left(x\right)$, and the non-linear mapping through stacked convolutional layers is represented as $F\left(x, W\right)$, in the context of residual networks, each residual module directly transmits input to output through shortcut connections. Consequently, for the forward pass of a residual module:
\begin{equation}
  x^{l+1}=f\left(F\left(x^l, W^l\right)+h\left(x^l\right)\right)\label{eq1}  
\end{equation}

If both $h$ and $f$ are set as identity mappings, we have:
\begin{equation}
x^{l+1}=F\left(x^l, W^l\right)+x^l
\end{equation}

Iterating further:
\begin{equation}
x^L=\sum_{i=1}^{L-1} F\left(x^i, W^i\right)+x^l
\end{equation}

For any two residual modules, the input features of shallow modules can be directly propagated backward, while the input features of deep modules are the accumulation of outputs from all preceding shallow modules. This highlights the characteristic of residual networks in promoting feature reuse and integration. Given that deep networks tend to converge towards an identity mapping ($H\left(x\right)=x$) as the optimal solution, the residual mapping is set to zero ($F\left(x\right)=0$) when needed, making the optimization of residual mappings easier than original non-linear mappings.
\begin{align}
\delta^l&=\frac{\partial E}{\partial x^l}\notag\\
        &=\frac{\partial E}{\partial x^L} \frac{\partial x^L}{\partial x^l}\notag\\
        &=\delta^L\left(1+\frac{\partial}{\partial x^l} \sum_{i=1}^{L-1} F\left(x^i, W^i\right)\right)\label{eq4} 
\end{align}

Similar to the forward pass, gradient backpropagation exhibits the same characteristics, which is shown as equation~\ref{eq4}. Furthermore, as gradients can propagate directly backward, even if weights are small, they do not tend to zero, alleviating the problem of gradient vanishing.

\begin{figure}[h]
\centering
\includegraphics[width=3in]{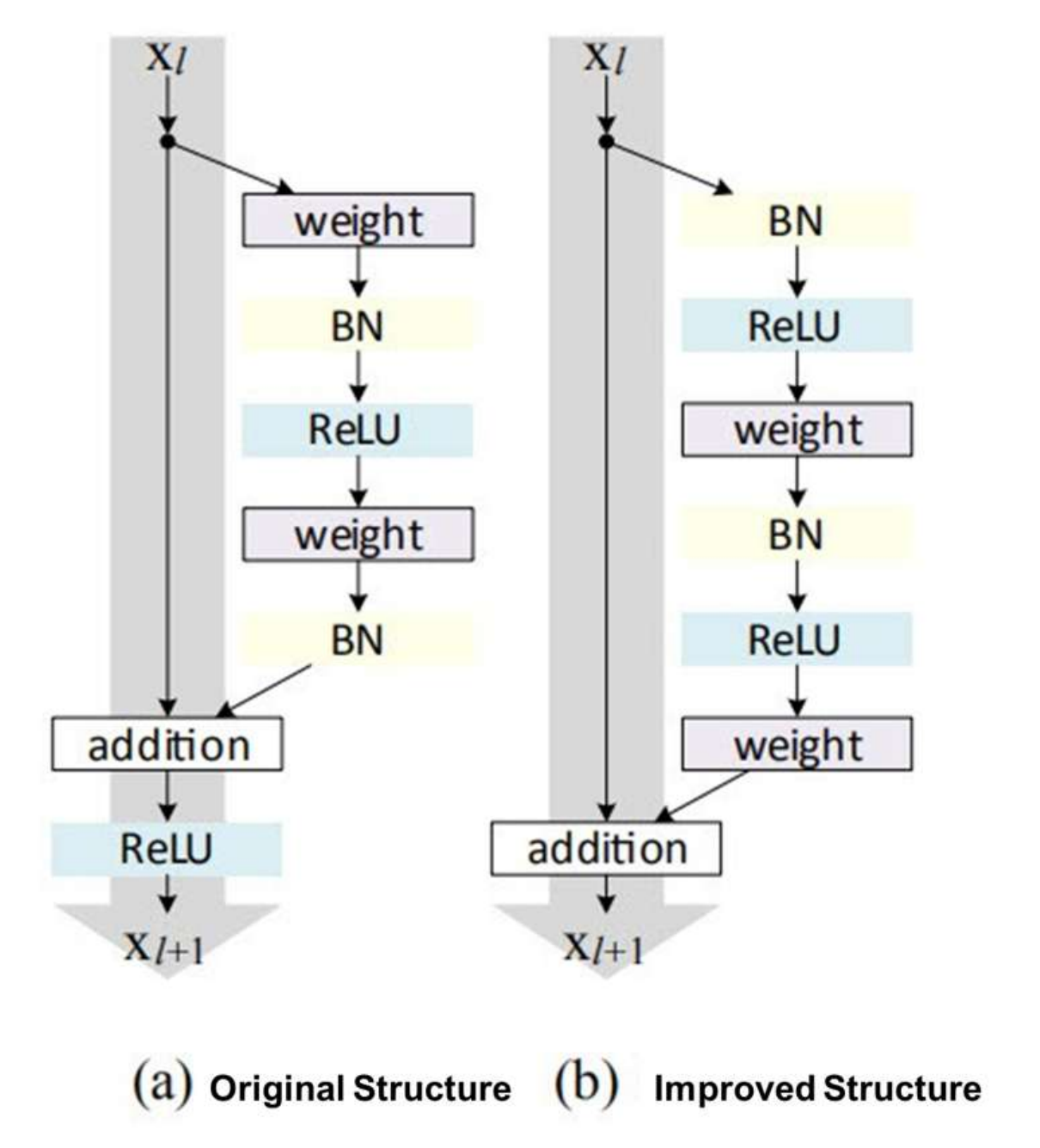} %
\caption{Improved residual module structure \cite{11}.}
\label{fig_2}
\end{figure}

Based on the aforementioned derivation, introducing any operation along the skip connection path accumulates effects as the network deepens, hindering information flow and making training difficult. To address this, Kaiming He et al. \cite{11} improved the structure of residual modules, as shown in  fig.~\ref{fig_2}.. This study adopted the enhanced residual module structure.

To further investigate the impact of skip connections on low-resolution image recognition, this paper conducted experiments and analysis on various forms of residual modules. It is essential to determine the fundamental dimensions of residual modules. The recognition accuracy of neural network models largely depends on the network's depth (number of layers) and width (number of channels). As depth and width increase, the network's complexity, representational power, and fitting capacity also increase. This paper introduces three fundamental dimensions for the studied problem: depth ($d$), width ($w$), and interlinks($i$).

\begin{figure}[h]
\centering
\includegraphics[width=3in]{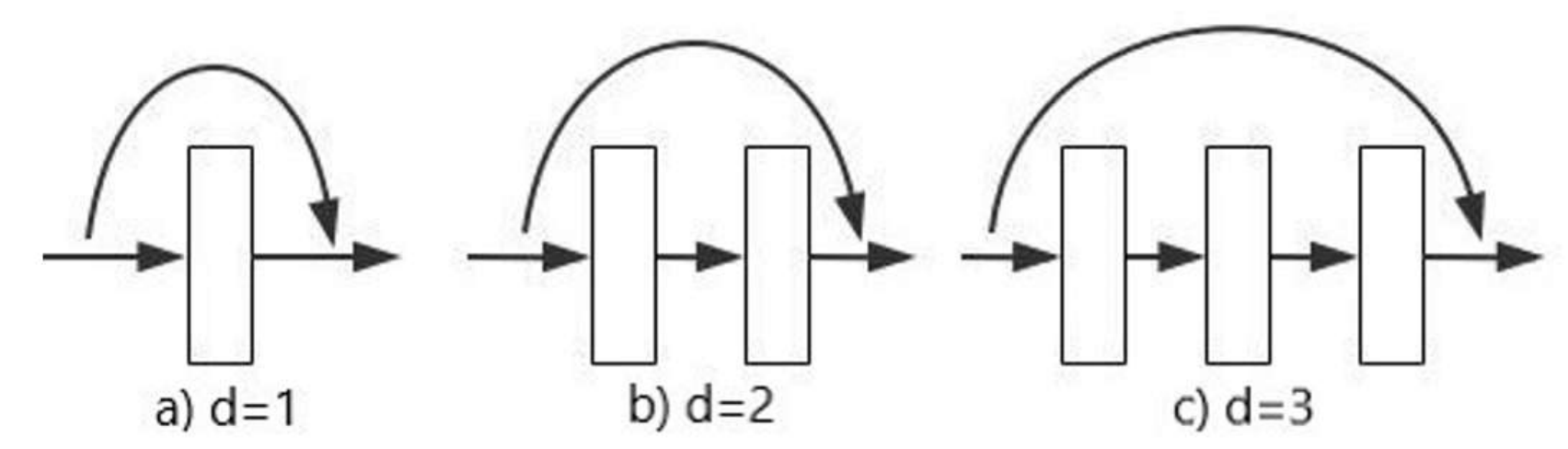} %
\caption{Residual module structures with different depths.}
\label{fig_3}
\end{figure}

Depth ($d$) refers to the number of convolutional layers within a single residual module. Depth primarily influences the density and span of skip connections. With increasing depth $d$, the number of skip connections decreases while the span increases. In other words, the degree of feature reuse decreases, and the differences in feature fusion increase. Different depths of residual module structures are shown in fig.~\ref{fig_3}.

Width ($w$) refers to a multiplicative factor for the number of channels in each convolutional layer within a residual module. Regarding feature extraction, width ($w$) represents the number of convolutional filters. As the number of filters increases for each layer, more features are extracted, and feature fusion during skip connections becomes more comprehensive. With an increase in width ($w$), the network's parameter count and computational requirements grow quadratically, but computational efficiency improves (favorable for GPU parallel processing).

\begin{figure}[h]
\centering
\includegraphics[width=3in]{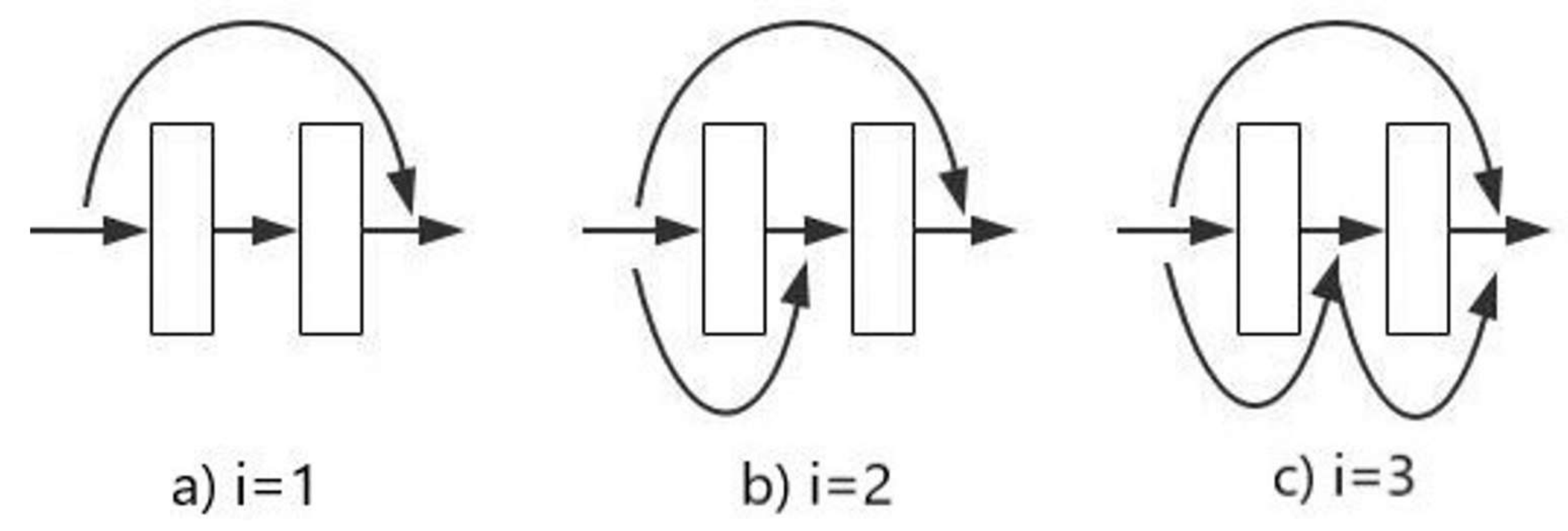} %
\caption{Residual module structures with different numbers of interlinks.}
\label{fig_4}
\end{figure}

Interlinks ($i$) refers to the number of skip connections within a single residual module. Residual module structures with different connectivity levels are shown in fig.~\ref{fig_4}. Increasing interlinks ($i$) does not introduce additional parameters and incurs relatively small computational overhead. With a fixed total network depth ($L$), and identical depths ($d$) and widths ($w$), more interlinks(i) leads to denser skip connections and greater feature reuse and fusion.

For clarity, the following symbols are adopted in the subsequent discussion: $rL-d-w-i$ represents a residual network with a total of $L$ layers, where an individual residual module has depth, width, and density denoted as $d$, $w$, and $i$, respectively (plain networks without inter-layer skip connections are denoted as plainnets).

\section{Recognition Algorithm for Low-Resolution Images Based on Dual-Branch Residual Networks}

While residual modules are advantageous for extracting discriminative intrinsic features from low-resolution images, a single residual network still struggles to fit the low-resolution image dataset, limiting its recognition performance. In order to further improve the recognition accuracy of low-resolution images, we propose a dual-branch residual network model, building upon the previous research. This model leverages the Common Feature Subspace Algorithm to introduce high-resolution image features during the training process of the low-resolution network. This guidance aids the network in converging in the correct direction. Given the similarity between the Common Feature Subspace Algorithm and knowledge distillation, we further combine knowledge distillation to reduce network parameters and computational complexity, making it more suitable for practical applications in intelligent vehicles.
\subsection{Common Feature Subspace and Knowledge Distillation}
The Common Feature Subspace Algorithm maps high and low-resolution images to the same subspace by minimizing the distance between corresponding features of high and low-resolution images, facilitating matching and recognition. In this context, the dual-branch network model utilizes HR (High-Resolution) and LR (Low-Resolution) networks to extract high and low-resolution features separately. Assuming the input high and low-resolution images are represented as XH and XL, and the mapping functions for HR and LR networks are denoted as $F_H$ and $F_H$, the objective function for the subspace algorithm can be expressed as:
\begin{equation}
J\left(F_H, F_L\right)=\sum_{i=1}^m\left\|F_H\left(x_i^{H R}\right)-F_L\left(x_i^{L R}\right)\right\|^2 
\end{equation}

To reduce model size and inference time, dual-branch network models are usually trained in a staged manner, following these steps:

\subsubsection*{\bf Step 1}
Train the HR network using a high-resolution image dataset (selecting cross-entropy or other appropriate loss functions based on the dataset characteristics).

\subsubsection*{\bf Step 2}
Keep the parameters of the HR network fixed and use high-resolution images as input to obtain high-resolution features. Train the LR network using these high-resolution features along with labels (employing a joint loss function that combines cross-entropy, Mean Squared Error (MSE), or other loss functions).

During the backward propagation of the LR network, the gradient of the objective function can be expressed as:
\begin{equation}
\delta=\frac{\partial J\left(F_H, F_L\right)}{\partial F_L\left(x_i^{L R}\right)}=\sum_{i=1}^m 2\left(F_H\left(x_i^{H R}\right)-F_L\left(x_i^{L R}\right)\right)
\end{equation}

The training process described above can be considered as knowledge transfer from the HR network to the LR network, allowing the LR network to learn mapping relationships from the high-resolution image dataset. High-resolution features obtained from the HR network contain more critical information than singular labels, aiding the LR network in converging in the right direction.

\subsection{Knowledge Distillation}

Knowledge distillation \cite{12} is a model compression method that transforms a structurally complex and highly accurate teacher network into a structurally simplified student network through "distillation," reducing model size and improving inference speed. As previously discussed, a model's expressive capacity is positively correlated with its complexity. With increased depth and channel count, the model acquires more knowledge from the dataset and performs better. However, in limited datasets, as the model's complexity continues to grow, improvements in recognition accuracy tend to plateau. In other words, overly complex models carry redundant parameters. To mitigate this redundancy, knowledge distillation employs a teacher network with a large parameter count and complex structure to fit a massive dataset during training, enhancing recognition accuracy. During deployment, a student network with a reduced parameter count and computational complexity is used for accelerated inference.
\begin{figure}[h]
\centering
\includegraphics[width=3in]{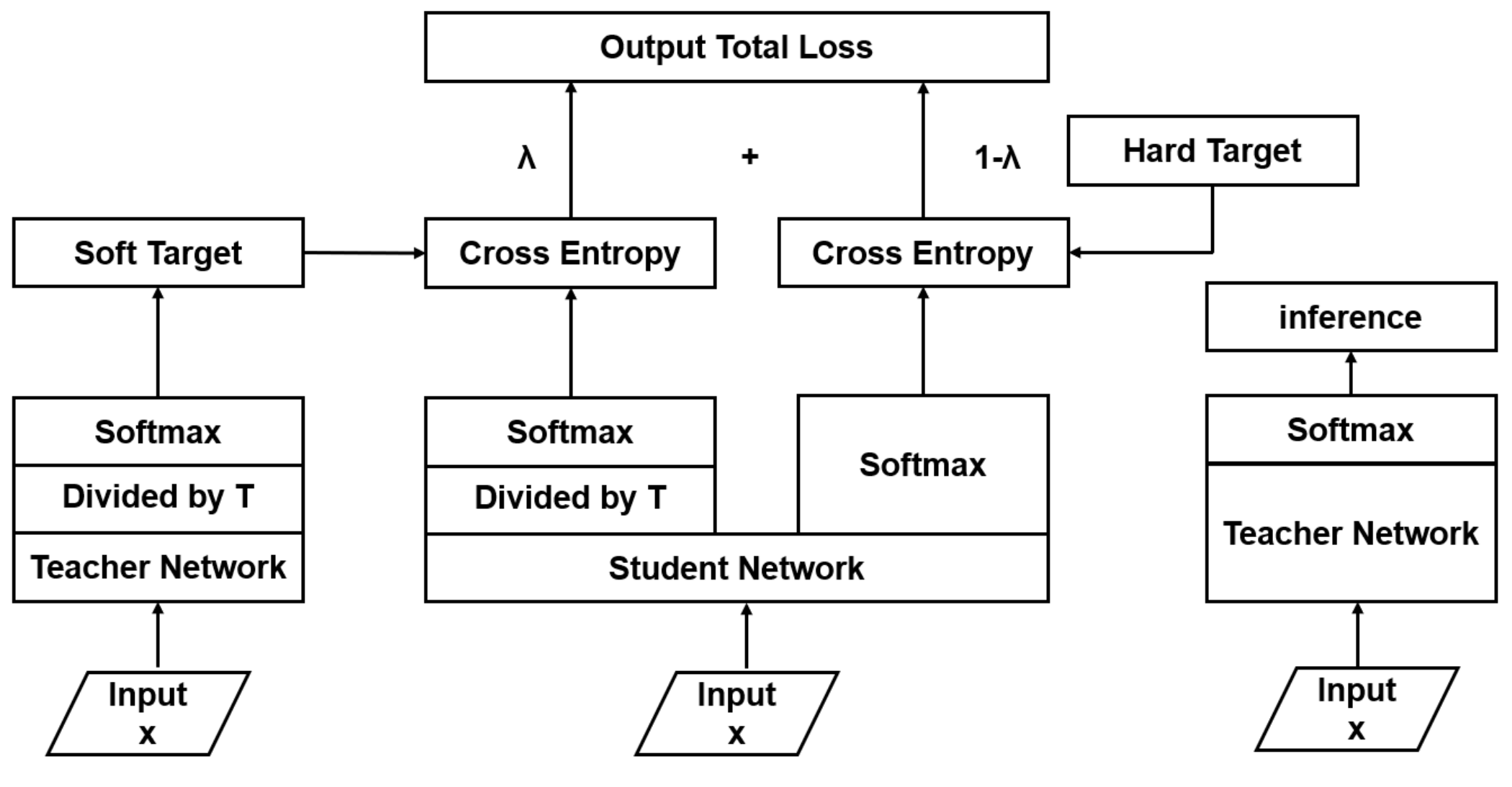} %
\caption{Process of Knowledge Distillation \cite{12}.}
\label{fig_5}
\end{figure}

The knowledge distillation process is depicted in fig.~\ref{fig_5} and consists of the following steps:

\subsubsection*{\bf Step 1}
Train the teacher network using the dataset (employing cross-entropy loss).

\subsubsection*{\bf Step 2}
Keep the teacher network parameters fixed and use the teacher network's outputs as soft targets to guide the training of the student network. The student network's loss function comprises two parts:

\begin{list}{}{}
\item{(1) Soft targets are obtained by dividing the feature vectors output by the teacher network's fully connected layer by a temperature parameter T and then passing them through a softmax layer. These soft targets and the predicted probability distribution from the student network's fully connected layer, also divided by the same temperature parameter T and passed through a softmax layer, constitute part of the loss function. } 
\item{(2) The other part of the loss function is the cross-entropy between the directly output predictions of the student network and the labels (also known as hard targets). }
\end{list}

\subsubsection*{\bf Step 3}
Infer with the trained student network.

The key to knowledge distillation is introducing soft targets to guide the training of the student network. Soft targets are probability distributions obtained from the outputs of the trained teacher network and contain more critical information than one-hot encoded labels. The temperature parameter $T$ is introduced to soften the probability distribution. A larger $T$ leads to more uniform probabilities among different classes. However, it can also amplify the influence of negative labels, increasing the proportion of effective information while potentially introducing noise that may hinder the convergence of the smaller model. Haitong Li \cite{13} conducted research on the temperature parameter and the weight of the soft target loss function, finding that when the weight is significant, the temperature parameter must be increased simultaneously to soften the probability distribution and achieve better training results. Additionally, knowledge distillation acts as a form of regularization, enhancing model generalization. As a result, larger learning rates and smaller datasets can be used.

\subsection{Intermediate Features and Attention Maps}

Overall, the Common Feature Subspace Algorithm and knowledge distillation share a high degree of similarity in terms of knowledge transfer. In the Common Feature Subspace Algorithm, the training process of high and low-resolution networks primarily focuses on knowledge transfer for different recognition tasks. In contrast, knowledge distillation trains teacher and student networks, focusing on knowledge transfer between models of different sizes. In essence, for individual learning processes, knowledge transfer is feasible between different individuals for different learning tasks. The transfer of knowledge must satisfy two conditions:

\begin{list}{}{}
\item{(1) Similarity: there must be sufficient similarity between learning tasks or models to ensure model convergence. } 
\item{(2) From larger to smaller: knowledge can only be transferred from the larger side to the smaller side, resulting in knowledge compression.}
\end{list}

To further enhance the effectiveness of knowledge transfer, literature \cite{14}\cite{15}\cite{16} has proposed using intermediate-layer features to guide the convergence of the LR network (or student network) for the Common Feature Subspace Algorithm and knowledge distillation. Two typical model structures are illustrated in the figure. Using intermediate features to construct the loss function helps fully utilize the valuable information contained in the intermediate layers and has a certain regularization effect. However, employing too many intermediate layer features can lead to a more complex loss function and training difficulties, potentially hindering network convergence. Therefore, it is essential to ensure the similarity of intermediate layer features and select intermediate layers to constitute the loss function reasonably.

Attention mechanisms select the most crucial information for the current task from a vast amount of information. Analyzing the differences in attention distribution among intermediate features of different objects can indicate their similarity. Attention maps can be divided into two types: the first is based on activations, reflecting the importance of neurons in intermediate layers through the magnitude of their activation values. The second type is gradient-based, indicating the sensitivity of a neuron to the input through the size of the gradient of that neuron. Both types of attention maps can reveal the spatial positions that the network "focuses on" in input images. For ease of setting up the loss function, the first type is used here. Its specific definition is as follows:

\begin{equation}
F_{\text {sum }}^p(x)=\frac{1}{D} \sum_{i=1}^D\left|x_i\right|^p \label{eq7}
\end{equation}

Where $x$ represents the feature map output from intermediate layers, $D$ is the number of channels in the feature map, and $p$ is the exponent (set to 2 here). In essence, the attention map is a two-dimensional plane obtained by averaging the $p$-th powers of the values of each channel in the feature map.

To effectively utilize intermediate layer features and avoid training difficulties that may arise from overly complex loss functions, it is necessary to measure the difference between intermediate layer features of HR and LR networks. Subsequently, the weight of the loss function can be determined. Assuming that HR and LR networks trained with high and low-resolution images, respectively, can reflect the differences in intermediate layers of the two branches of the dual-branch network during training, we directly use the "attention loss" of HR and LR networks to measure this difference, defined as follows:

\begin{equation}
E_{A T}\left(b l o c k_j\right)=\frac{1}{m} \sum_{i=1}^m \frac{1}{q}\left\|\frac{Q_{i j}^{H R}}{\left\|Q_{i j}^{H R}\right\|_2}-\frac{Q_{i j}^{L R}}{\left\|Q_{i j}^{L R}\right\|_2}\right\|_2 \label{eq8}
\end{equation}

Where $Q$ is the vector obtained by flattening the attention map, $i$ and $j$ represent the $i$-th input image and the $j$-th block, and $m$ is the batch size (set to 128 here). To make the loss values from attention maps of different sizes comparable, $Q$ in the equation.~\ref{eq8} is L2 normalized and divided by its vector length $q$.

\subsection{Dual-Branch Residual Network}

\begin{figure}[h]
\centering
\includegraphics[width=3in]{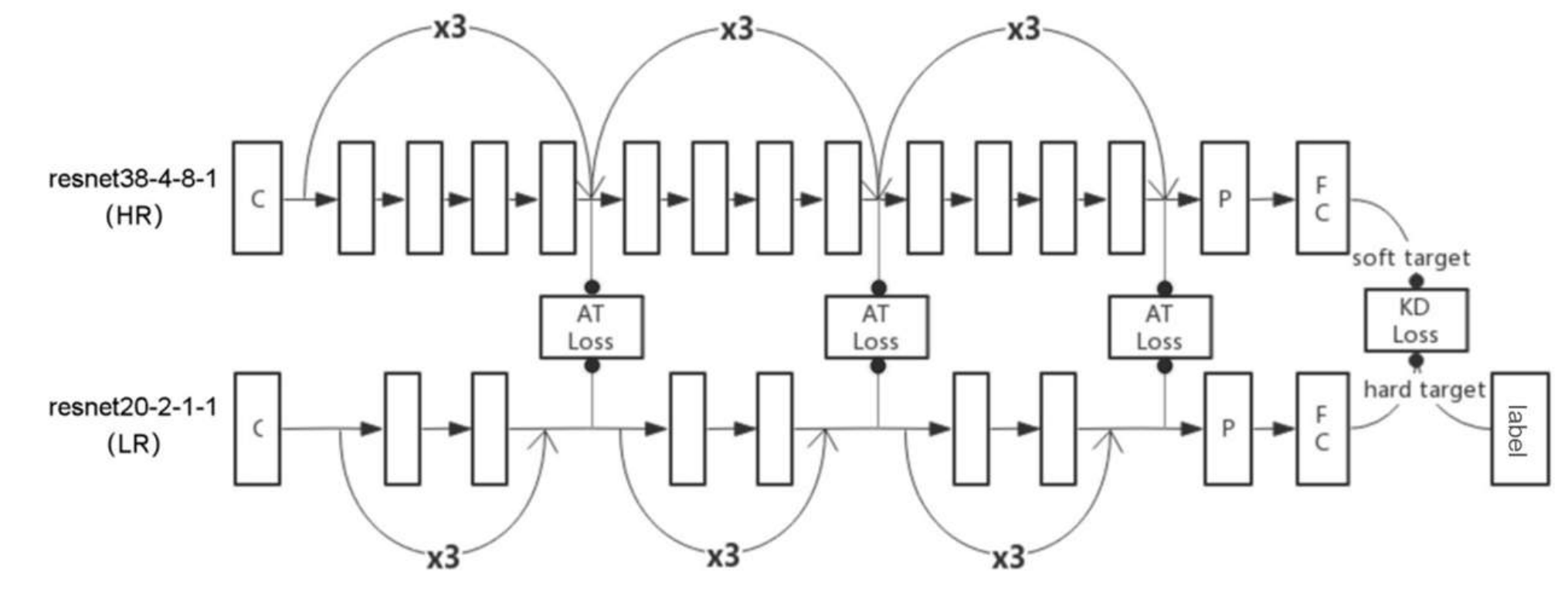} %
\caption{The Dual-Branch Residual Network Model Architecture.}
\label{fig_6}
\end{figure}

Combining the Common Feature Subspace Algorithm and knowledge distillation, this section proposes a dual-branch residual network model to address the recognition problem of low-resolution images, as illustrated in the fig.~\ref{fig_6}. The entire model consists of two branches: an HR network and an LR network. The HR network employs a complex network with a deep layer and a large width, $r38-4-8-1$, while the LR network uses a "lightweight" network, $r20-2-1-1$. The features output by various intermediate layers and fully connected layers of the HR network are used to guide the training of the LR network, enabling knowledge transfer from high-resolution to low-resolution and from a large network to a small network.

During HR network training, a standard softmax cross-entropy loss function augmented with regularization terms is used to enhance the network's ability to extract features from high-resolution images. Training of the LR network employs a joint loss function composed of the following three parts:

\subsubsection*{\bf Knowledge distillation loss}

This loss comprises two weighted components, hard targets (labels) and soft targets (probability distribution obtained by dividing the output from the HR network's fully connected layer by the temperature parameter T and passing it through a softmax layer). It is used to improve the recognition accuracy of the LR network for low-resolution images. The expression is as follows:

\begin{align}
E_{K D h} & =-\frac{1}{m} \sum_{i=1}^m \sum_{j=1}^n y_i(j) \log \left(\frac{e^{x_i(j)}}{\sum_{j=1}^n e^{x_i(j)}}\right)\\
q_i(j) & =\frac{e^{x_i(j) / T}}{\sum_{j=1}^n e^{x_i(j) / T}}\\
E_{K D s} & =-\frac{1}{m} \sum_{i=1}^m \sum_{j=1}^n q_i^h(j) \log \left(q_i^l(j)\right)\\
E_{K D} & =(1-\alpha) E_{K D h}+\alpha T^2 E_{K D s}
\end{align}

Where $T$ is the temperature parameter, and $\alpha$ is the loss function weight.

\subsubsection*{\bf Attention loss}
This loss is obtained by weighting the attention losses from block1, block2, and block3 outputs. It guides the LR network to "focus" on locations containing crucial feature information. The expression is as follows:

\begin{align}
E_{A T}\left(\text { block }_j\right) & =\frac{1}{m} \sum_{i=1}^m \frac{1}{q}\left\|\frac{Q_{i j}^{H R}}{\left\|Q_{i j}^{H R}\right\|_2}-\frac{Q_{i j}^{L R}}{\left\|Q_{i j}^{L R}\right\|_2}\right\|_2 \\
E_{A T} & =\frac{\beta}{2}\omega_1 E_{A T}\left(\text { block }_1\right)\notag\\
& +\frac{\beta}{2}\omega_2 E_{A T}\left(\text { block }_2\right)\notag\\
& +\frac{\beta}{2}\omega_3 E_{A T}\left(\text { block }_3\right)
\end{align}
Where, $\beta$, $\omega_1$ , $\omega_2$ , $\omega_3$ are loss function weights.

 \subsubsection*{\bf Regularization loss}
This loss is a penalty term consisting of the square values of the LR network's weights, preventing overfitting. The expression is as follows:

\begin{equation}
E_{R E G}=\frac{\lambda}{2}\|W\|_2^2
\end{equation}

Where $\lambda$ is the loss function weight, also known as the weight decay coefficient.

The joint loss function is as follows:

\begin{equation}
E=E_{K D}+E_{A T}+E_{R E G}
\end{equation}

\section{Experiments and Results}
\subsection{Dataset and experimental setup}

This paper uses CIFAR-10 data set for image recognition experiments. CIFAR-10 is a small data set commonly used in image classification problems. The data set includes a training set with 50,000 images and a test set with 10000 images. During the experiment, the data set was divided into three groups of images of different resolutions: 32 × 32 pixels, 16 × 16 pixels, and 8 × 8 pixels. The first type preserves the original image size of the dataset; The latter two types obtain low-resolution images by image downsampling, and add random noise to the images to simulate the interference in the actual scene. The above high and low-resolution images are enhanced by random clipping and random horizontal flipping in the training process to expand the sample size of the training set.

The software environment of the experiment is implemented by the TensorFlow framework of deep learning. All experiments are implemented by NVIDIA Tesla T4, P100, V100 and other GPUs provided by Google Colaboratory platform. The training method of the experimental model adopts the random gradient descent method, in which the weight decay is set to 0.0001, the momentum is set to 0.9, the batch size of a sample is set to 128, and the number of training rounds is set to 64000. Using the learning rate of piecewise constant attenuation, the initial learning rate is set to 0.1, which decreases to 0.01 and 0.001 respectively with the number of training rounds reaching 32000 and 48000.

\subsection{Comparison of ResNet with Other Models}
\begin{figure}[h]
\centering
\includegraphics[width=3in]{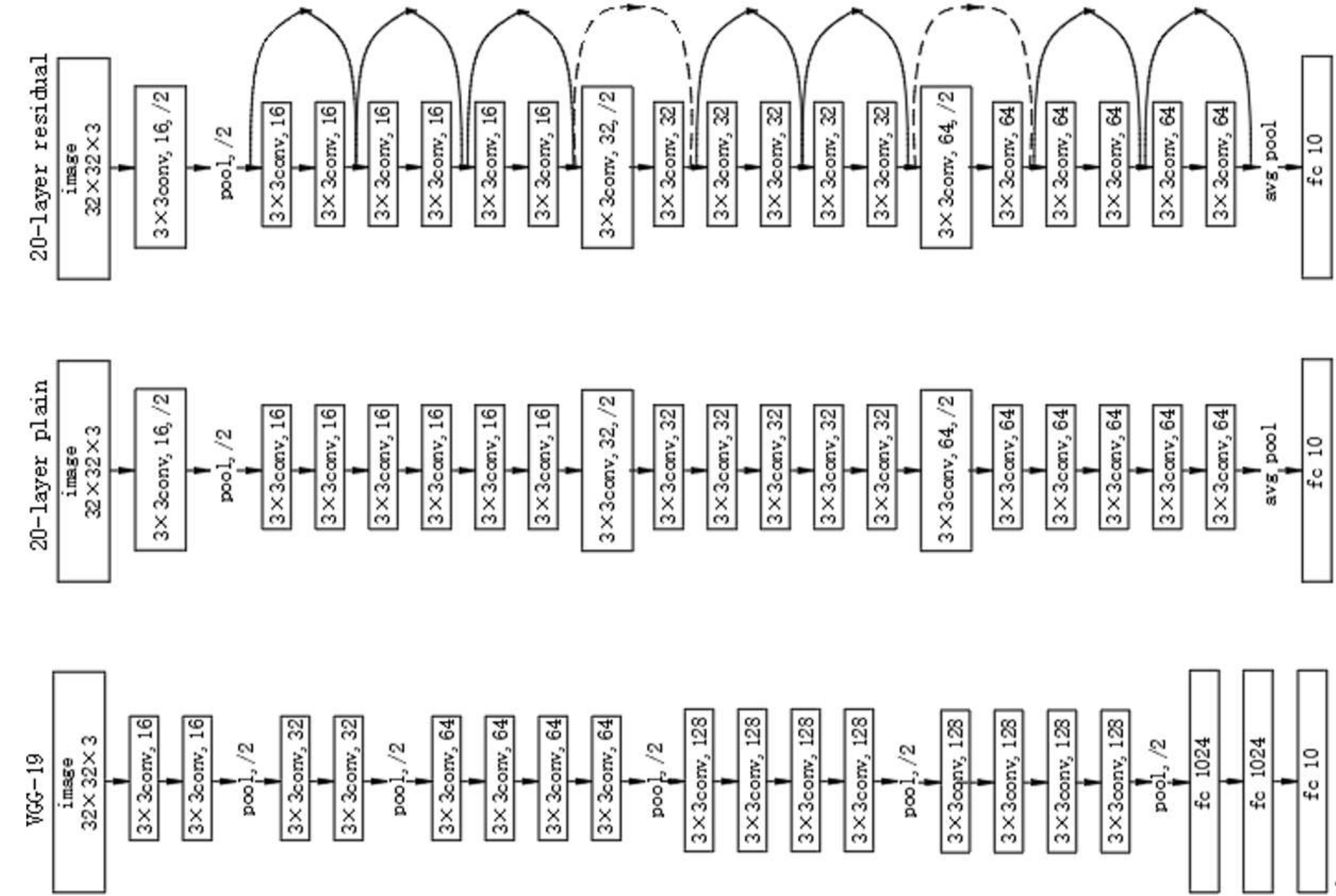} %
\caption{Three different network structures.}
\label{fig_7}
\end{figure}

In this experiment, VGG19, plainnet20 and ResNet20-2-1-1 with similar layers are selected for comparison. The structures of these three networks are shown in the fig.~\ref{fig_7}. In order to express concisely and clearly, the experimental data and charts in the following text use p for plainnet and r for ResNet. (for example, p20=plainnet20 or r38-2-1-1= ResNet38-2-1-1)

The experimental results of the above three network models for CIFAR-10 original resolution image recognition are shown in Table~\ref{tab1}. The accuracy of plainnet20 and ResNet20 models for data set original resolution image recognition is significantly higher than that of VGG19 model. In addition, the more important point is that the latter two models have fewer filters and lower complexity than the VGG19 model. In terms of floating-point operands (FLOPs), the calculation amount of the benchmark model with 20 layers is only 3.42M, which is 13\% of the calculation amount of VGG19 (26.47 M FLOPs).

\begin{table}[h]
\caption{Recognition accuracy of three network models for raw images in the CIFAR-10 dataset\label{tab1}}
\centering
\begin{tabular}{l|l}
\hline
\begin{tabular}[c]{@{}l@{}}Network\\ model\end{tabular} & \begin{tabular}[c]{@{}l@{}}Accuracy(\%)\\ for 32$\times$32\end{tabular} \\ \hline
VGG19                                                   & 88.27                                                              \\
plainnet20                                              & 90.10                                                              \\
ResNet20                                                & 91.34                                                              \\ \hline
\end{tabular}
\end{table}

In order to verify the effect of the residual module on low-resolution image recognition, another group of experiments further compared the effects of plainnet20 and ResNet20 models on 16×16 and 8×8 resolution image recognition. The experimental results are shown in Table.~\ref{tab2} below.

\begin{table}[h]
\caption{The impact of the residual module in neural networks on the recognition of low-resolution images (16×16, 8×8) \label{tab2}}
\centering
\begin{tabular}{@{}l|l|l@{}}
\hline
\begin{tabular}[c]{@{}l@{}}Network\\ model\end{tabular} &
  \begin{tabular}[c]{@{}l@{}}Accuracy(\%)\\ for 16 × 16\end{tabular} &
  \begin{tabular}[c]{@{}l@{}}Accuracy(\%)\\ for 8 × 8\end{tabular} \\ \hline
plainnet20 &
  84.49 &
  68.74 \\
ResNet20 &
  85.73 &
  69.74 \\ \hline
\end{tabular}
\end{table}

Whether it is 8×8 resolution image, or 16×16 resolution images, the accuracy of ResNet is significantly higher than that of plainnet with the same number of layers but without residual module. Therefore, we decided that the subsequent experiments would be based on ResNet model with skip connections.

\subsection{Residual network experiment}
In order to further study the impact of Skip Connections on low-resolution image recognition, this paper tests and analyzes the performance of different forms of residual module from the three basic dimensions of residual network: depth $d$, width $w$ and interlinks(correlation degree) $i$. Finally, we determined the structure and parameters of residual network used in this paper.

\begin{table}[h]
\caption{basic structure of residual network \label{tab3}}
\centering
\begin{tabular}{@{}l|l|l@{}}
\hline
Layers & Output size & ResNetL-d-w-i              \\ \hline
conv   & 32 $\times$ 32 & $[3 \times 3,16]$                          \\
block1 & 32 $\times$ 32 & ${\left[\begin{array}{l}
3 \times 3,16 \times w \\
3 \times 3,16 \times w
\end{array}\right] \times N}$                           \\
block2 & 16 $\times$ 16         & ${\left[\begin{array}{l}
3 \times 3,32 \times w \\
3 \times 3,32 \times w
\end{array}\right] \times N}$                           \\
block3 & 8 $\times$ 8         & ${\left[\begin{array}{l}
3 \times 3,64 \times w \\
3 \times 3,64 \times w
\end{array}\right] \times N}$                          \\ \hline
       & 1 $\times$ 1         & avg-pool, 10-d fc, softmax \\ \hline
\end{tabular}
\end{table}

To compare the effects produced by different forms of residual modules, the basic structure of the residual network should remain consistent in the experiment, as shown in Table.~\ref{tab3} The network consists of an input convolutional layer, three sets of residual blocks (blcok1, blcok2, blcok3), a global average pooling layer, and a final fully connected layer. Each block consists of $N$ residual blocks, with the number of channels doubling from block1 to block3. To achieve dimensional reduction, the input convolutional layers of block2 and block3 use 3×3 convolution kernels with a stride of 2, and the corresponding skip connections use 1×1 convolution kernels with a stride of 2 to ensure dimensional matching. The residual module adopts the structure shown in fig.~\ref{fig_2}(b), arranged in the order of BN-ReLU-weight, which can accelerate the convergence of the network and improve the generalization performance of the model \cite{17}.

\subsubsection*{\bf 1. Performance of ResNet with different layers ($L$) on recognizing images.}
By adjusting the number of network layers, the recognition accuracy of ResNet for images with different resolutions is shown in Table.~\ref{tab4}.

\begin{table}[h]
\caption{Accuracy of ResNet model with different layers for image recognition\label{tab4}}
\centering
\begin{tabular}{@{}l|l|l@{}}
\hline
\begin{tabular}[c]{@{}l@{}}Network\\ model\end{tabular} &
  \begin{tabular}[c]{@{}l@{}}Accuracy(\%)\\ for 32×32\end{tabular} &
  \begin{tabular}[c]{@{}l@{}}Accuracy(\%)\\ for 8×8\end{tabular} \\ \hline
r20-2-2-1  & 91.22 & 69.74 \\
r38-2-1-1  & 92.15 & 70.18 \\
r110-2-1-1 & 93.10 & 71.34 \\ \hline
\end{tabular}
\end{table}

The ResNet model using the residual module shows an increasing trend in accuracy on the original dataset as the number of layers gradually deepens; Low resolution (8 × 8) In the image, r38 is slightly better than r20, and network degradation is suppressed. r110 clearly fits the relationship of increasing recognition accuracy. However, the r110 network also brings a significant computational load problem as the number of layers increases, making the training process relatively slow. Compared to the small increase in accuracy, the exponential increase in training time is not conducive to the experimental progress of subsequent adjustments to other parameters of the network model. 

Here, the number of ResNet model layers in subsequent experiments is selected as 38, mainly for the following reasons: 
\begin{list}{}{}
\item{1. Compared to r20 and r110, the r38 model has moderate computational load and accuracy; } 
\item{2. The low number of network layers in r20 resulted in a narrow range of parameter adjustments in the residual module for subsequent experiments, while r38 allows for appropriate changes to the "convolutional layer number of a single residual module" (referred to as depth $d$ in this paper).}
\end{list}
\par
 The ResNet model with 38 layers is selected for the high-resolution network, but in order to further reduce the network complexity at the deployment end, the number of layers of the low-resolution network is selected as 20.

\subsubsection*{\bf 2. Performance of ResNet with different depths ($d$) on recognizing images}

In order to compare the influence of residual modules with different depths on the recognition effect, we must ensure that the complexity of the network is the same, that is, the total parameters are unchanged. Therefore, while increasing the depth $d$, the number of residual modules n should be reduced so that the total number of layers of the network $L=N × D$ remains unchanged (the first convolution layer and the last full connection layer are not included here)

\begin{table}[h]
\caption{Accuracy of ResNet model with different depths for image recognition\label{tab5}}
\centering
\begin{tabular}{@{}l|l|l@{}}
\toprule
\begin{tabular}[c]{@{}l@{}}Network\\ model\end{tabular} &
  \begin{tabular}[c]{@{}l@{}}Accuracy(\%)\\ for 32×32\end{tabular} &
  \begin{tabular}[c]{@{}l@{}}Accuracy(\%)\\ for 8×8\end{tabular} \\ \midrule
r38-1-1-1 & 91.50 & 69.10 \\
r38-2-1-1 & 92.21 & 70.18 \\
r38-3-1-1 & 92.14 & 71.34 \\
r38-4-1-1 & 92.10 & 70.46 \\
r38-6-1-1 & 92.15 & 69.76 \\ \bottomrule                                         
\end{tabular}
\end{table}

The experimental results (Table.~\ref{tab5} show that the change of R38 network model depth has little effect on the accuracy of the original image of the data set. When the depth is 1, each residual module contains only one convolution layer, so that the number of residual modules is the largest, and the reuse of features is too much, which has a negative impact on the results; After the depth is greater than or equal to 2, the increase of depth means that the span of skip connection is increased, the degree of feature reuse is reduced, and the accuracy rate is basically maintained at a high level.
The effect of depth change on the low-resolution image is basically consistent with the above experimental results of the original image, but when the low-resolution image has a small number of features, the model with a maximum depth of 6 shows some drawbacks in the experimental results. Its large skip connection span makes the degree of feature reuse very low. For the low-resolution image input itself, the number of features is quite small, and the difference of feature fusion is very large, resulting in the reduction of accuracy.

Considering comprehensively, the r38 model with depth of 4 is the best in the experimental results, which can be used as a follow-up experiment.

\subsubsection*{\bf 3. Performance of ResNet with different widths ($w$) on recognizing images}

This paper refers to the approach of reference \cite{18} and conducts comparative experiments on different combinations of L and w. The experimental results are shown in Table.~\ref{tab6}. Respectively adjust the width of the r20 and r38 network models for experiments, and the results of the original images in the dataset are consistent with the conclusions of the literature. Whether it is a low-level r20 network or a higher-level r38 network, as the number of channels in each convolutional layer increases, the output accuracy significantly improves. At this time, the number of parameters in the network increases correspondingly with the amount of computation required, and although the increase in the number of channels is more suitable for the parallel computing characteristics of GPUs to a certain extent, the model training period is longer than in previous parameter adjustment experiments.

\begin{table}[h]
\caption{Accuracy of ResNet model with different widths for image recognition.\label{tab6}}
\centering
\begin{tabular}{@{}l|l|l@{}}
\hline 
\begin{tabular}[c]{@{}l@{}}Network\\ model\end{tabular} &
  \begin{tabular}[c]{@{}l@{}}Accuracy(\%)\\ for 32×32\end{tabular} &
  \begin{tabular}[c]{@{}l@{}}Accuracy(\%)\\ for 8×8\end{tabular} \\ \hline 
r20-2-1-1 & 91.22 & 69.74 \\
r20-2-2-1 & 92.92 & 71.48 \\
r20-2-4-1 & 93.83 & 73.38 \\
r20-2-8-1 & 94.32 & 73.80 \\
r38-4-1-1 & 92.09 & 70.46 \\
r38-4-2-1 & 93.54 & 72.46 \\
r38-4-4-1 & 94.49 & 73.55 \\
r38-4-8-1 & 95.04 & 74.48 \\ \hline     
\end{tabular}
\end{table}

As for training on low-resolution datasets，the results show that for low-resolution inputs, a larger number of channels can achieve significantly higher accuracy than the original residual network ($w=1$). In addition, on low-resolution datasets, changing the width has a greater impact on the results than the previously discussed parameters. We decided to set the width of r32 to 8, which will be used for following experiments and the high-resolution network. As for the width of low-resolution ResNet20, we choose 1.

\subsubsection*{\bf 4. Performance of different numbers of interlink ($i$) on recognizing images}
Similar to the previous experiments, only the numbers of interlink ($i$) is changed in each network, and the results are shown in Table.~\ref{tab7}.

\begin{table}[h]
\caption{Accuracy of ResNet model with different number of interlinks for image recognition\label{tab7}}
\centering
\begin{tabular}{@{}l|l|l@{}}
\hline  
\begin{tabular}[c]{@{}l@{}}Network\\ model\end{tabular} &
  \begin{tabular}[c]{@{}l@{}}Accuracy(\%)\\ for 32×32\end{tabular} &
  \begin{tabular}[c]{@{}l@{}}Accuracy(\%)\\ for 8×8\end{tabular} \\ \hline  
r20-2-1-1 & 91.22 & 69.74 \\
r20-2-1-2 & 92.29 & 69.66 \\
r20-2-1-3 & 90.62 & 69.99 \\
r38-4-8-1 & 95.04 & 74.48 \\
r38-4-8-2 & 95.11 & 74.36 \\
r38-4-8-3 & 95.02 & 73.91 \\ \hline  
\end{tabular}
\end{table} 

Increasing the number of interlinks makes the skip connections become more intensive, and increases the reuse and fusion of features. At the same time, it will not increase the amount of parameters and calculations. It can be observed from the experimental results in Table.~\ref{tab7} that the change in the number of interlink $i$ has little effect on the accuracy. Too many interlinks will cause a slight decline in accuracy. As the skip connections become dense, too much feature reuse and fusion is not conducive to the training process. The number of interlinks of both high- and low-resolution network are selected as 1.  

\subsection{Attention loss test}
In this experiment, we conducted a visual analysis of intermediate layer features in different architectures of high and low-resolution networks, leveraging attention maps proposed by Sergey Zagoruyko et al. \cite{16}.

\begin{figure}[h]
\centering
\includegraphics[width=3in]{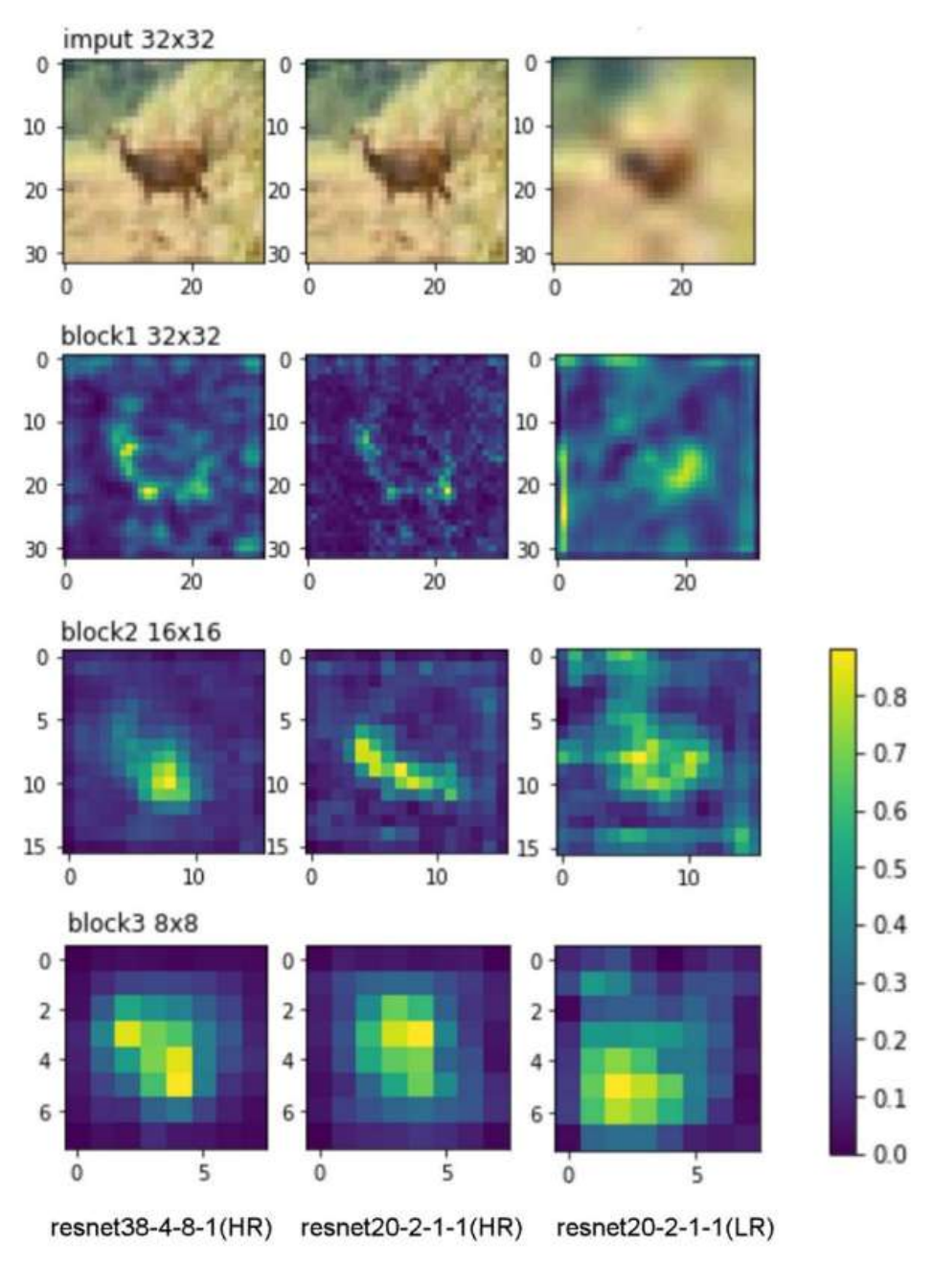} %
\caption{Visualization results of the attention map of the middle layer.}
\label{fig_8}
\end{figure}

We employed two different structures for our HR (High-Resolution) networks, namely ResNet38-4-8-1 and ResNet20-2-1-1, trained on a high-resolution image dataset. Conversely, the LR (Low-Resolution) network utilized the ResNet20-2-1-1 architecture and was directly trained on a low-resolution image dataset. Attention maps were visualized (shown as fig.~\ref{fig_8}, for the outputs of block1, block2, and block3, with image sizes of 32×32, 16×16, and 8×8, respectively.

Since input images were normalized, the values of each pixel in the attention maps ranged from 0 to 1, with pixel values represented by a color gradient from deep blue to yellow, as indicated by the color bar in the figures. As mentioned previously, attention maps reflect the concentration of the network's "attention." Pixels closer to yellow indicate higher levels of "attention."

Observing the attention maps from different layers, as the depth of the network increased, the semantic information in the images gradually became more abstract. High-level features became less distinguishable. For the shallow block1 outputs, the network's attention was primarily focused on recognizing the target (in this case, a deer), particularly on the three key areas containing crucial features: the head, body, and tail.

Comparing the attention maps of high and low-resolution networks, both HR networks accurately captured the key locations, while the LR network exhibited a lack of concentration in ‘attention’. Many irrelevant yellow points appeared at the edges of the attention map for block1 outputs in the LR network, leading to significant deviations in extracted high-level features compared to the HR networks. This suggests that the deficiencies in low-resolution images indeed hinder the feature extraction process.

\begin{table}[h]
\caption{Attention loss of each block output between HR and LR networks\label{tab8}}
\centering
\begin{tabular}{@{}l|l|l@{}}
\hline    
layers & \begin{tabular}[c]{@{}l@{}}ResNet38-4-8-1(HR)\\ ResNet20-2-1-1(HR)\end{tabular} & \begin{tabular}[c]{@{}l@{}}ResNet20-2-1-1(HR)\\ ResNet20-2-1-1(LR)\end{tabular} \\ \hline    
block1 & 0.003151 & 0.003156 \\
block2 & 0.001992 & 0.001690 \\
block3 & 0.001940 & 0.001262 \\
sum    & 0.007083 & 0.006108 \\ \hline                             
\end{tabular}
\end{table}

Subsequently, we calculated the attention loss distribution for each block output between the LR network and the two HR networks based on the "attention loss function" (equation~\ref{eq8}), with a batch size of 128. The results are shown as Table.~\ref{tab8}. In general, networks with the same structure exhibited smaller attention losses between HR and LR networks. Furthermore, as the depth increased, the attention losses for the corresponding block outputs gradually decreased. Among them, block1 from the shallow layers had the highest loss values, indicating the greatest difficulty in fitting.

\subsection{Training of Dual-Branch Residual Network}

In this experiment, we trained the proposed "Low-Resolution Image Recognition Algorithm based on the Dual-Branch Residual Network" using high and low-resolution images from the CIFAR-10 dataset and obtained the final results.

As described in the 'Dual-Branch Residual Network' section, this study introduced a Dual-Branch Residual Network model, integrating common feature subspace algorithms, knowledge distillation, intermediate features, and attention mechanisms. The joint loss function for training the low-resolution network includes knowledge distillation loss, attention loss, and regularization loss.

We initially trained the HR (High-Resolution) network using a high-resolution image dataset and saved the model parameters with the highest test accuracy. Subsequently, we artificially generated a low-resolution image dataset from the high-resolution image dataset. Specifically, we downsampled the images to the desired low resolution, followed by bicubic interpolation upscaling and the addition of Gaussian noise. We then replicated the HR network model while keeping its parameters unchanged. High and low-resolution images were separately input into the HR and LR (Low-Resolution) networks to obtain intermediate feature maps and fully connected layer output vectors. And then, we take the following steps to calculate loss functions:

\begin{list}{}{}
\item{1. The three-dimensional feature maps from the intermediate layers, namely block1, block2, and block3, were transformed into two-dimensional attention maps with sizes of 32 × 32, 16 × 16, and 8 × 8, respectively, following equation 14 to calculate the attention loss.} 
\item{2. The output vectors from the fully connected layers of both LR and HR networks, as well as the label vectors, all had a length of 10. Knowledge distillation loss was computed according to equations 9-12.}
\end{list}

The LR network was trained using a joint loss function comprising attention loss, knowledge distillation loss, and regularization loss. This is the end of the training process.

\begin{table}[h]
\caption{The accuracy of single-network and dual-branch network in recognizing low-resolution (8×8) images\label{tab9}}
\centering
\begin{tabular}{@{}l|l@{}}
\hline
\begin{tabular}[c]{@{}l@{}}Network\\ model\end{tabular} & \begin{tabular}[c]{@{}l@{}}Accuracy(\%)\\ 8 × 8\end{tabular} \\ \hline
r20-2-1-1(single)                                       & 71.07                                                    \\
r20-2-1-1(dual-branch)                                  & 72.15                                                    \\ \hline
\end{tabular}
\end{table}

The experiment still utilized the CIFAR-10 dataset. The weight parameter $\alpha$ for knowledge distillation loss was set to 0.9, temperature parameter $T$ was set to 4, attention loss weight parameter $\beta$ was set to 0.1, while $\omega_1$, $\omega_2$, and $\omega_3$ were chosen based on the values from Table.~\ref{tab8} (items with larger losses corresponded to smaller weights to reduce training difficulty). The regularization loss weight decay coefficient was set to 0.005. The experimental results are presented in Table.~\ref{tab9}.

\section{Conclusion and Discussion}

This paper aims to solve the problem of recognition of low-resolution images collected in the practical application of intelligent vehicles. We propose to use the residual network structure as the foundation, and incorporate the double branch network structure and feature subspace method for recognizing low-resolution images. This approach leads to the development of a low-resolution image recognition algorithm based on the double branch residual network. The algorithm provides an effective solution for promptly and accurately identifying the external environment of intelligent vehicles.
 
Firstly, we analyze the problem of feature extraction in traditional machine vision and typical neural network models for low-resolution images. When the above two methods recognize low-resolution images, they have large attenuation in accuracy and are not robust. The traditional target detection methods mainly include inter-frame difference method, optical flow method and so on. The most common method of moving object detection based on OpenCV - the inter-frame difference method \cite{19} is to take out the images of two consecutive frames in the video stream, and then subtract the corresponding matrix positions in mathematics to obtain the difference between the two frames of images, and then use the cv.findcontours function of OpenCV to obtain the position of the moving object. However, the disadvantage of this method is that the camera position should be absolutely fixed without any jitter, otherwise the environmental characteristics are also regarded as vehicle characteristics, which seriously interferes with the detection effect. The target detection of optical flow method \cite{20} corresponds each pixel of a frame image in the video stream to an optical flow vector. The optical flow method can detect whether there is a target passing in the scene, convert the image into a vector field to judge the target, and regard the background information as a uniform vector field. If there is any target passing through the monitored scene, a new vector field will be obtained. The vector field is different from the vector field of the previous frame, so it can judge whether there is a vehicle target passing at that time. However, the disadvantages of this method are also obvious. For example, the calculation of optical flow method needs to go through multi-cycle iterative calculation, which significantly increases the computational complexity and cannot achieve real-time detection and recognition. Therefore, the method based entirely on OpenCV has too many limitations and is difficult to be widely applied in practice. In terms of typical neural networks, scholars \cite{21} made in-depth research and improvements in the method of generating candidate regions. Through target analysis, they proposed to use the method based on visual saliency to obtain candidate regions, and then extract the features of candidate regions to reduce the dimension of data. However, this method uses the traditional CNN, which can not meet the real-time requirements and can only be used for offline identification. In order to achieve a certain balance between speed and accuracy, after a lot of research and exploration, scholars further proposed the target detection method using regression. YOLO\cite{22} (you only look once) is the most typical and widely used convolutional neural network based on regression.

Compared with the traditional feature extraction algorithm, convolutional neural network has the feature representation of adaptive learning training data, and does not need to manually design some features for specific problems, which can simplify the feature model and improve the efficiency at the same time. Secondly, the convolutional neural network is invariant to translation, rotation and other deformation to a certain extent, which can also improve image recognition accuracy. Therefore, we confirm that the deep neural network has high performance in feature extraction, and it can be applied to solve the problem of low-resolution image recognition.
 
The residual network (ResNet) in the deep network has the characteristics of feature reuse and fusion, and can adapt to low-resolution images. We compared it with VGG and plainnet network. ResNet model has the highest recognition accuracy for the cifar-10 dataset. Compared with the VGG19 model, it has fewer filters and lower complexity. In terms of floating-point operands (flops), the ResNet benchmark model with 20 layers has only 3.42m flops, which is 13\% of the VGG19 calculation (26.47m flops). So we decided to use ResNet as the neural network structure of this paper. Then, we carry out quantitative experiments on low-resolution image recognition around the model structure, number of layers and the basic dimensions of the residual module (depth $d$, width $w$ and correlation degree $i$). According to a large number of experimental data, we analyze the influence of various network parameters on the results and make the optimal choice.
 
Based on the residual module research, in order to further improve the recognition accuracy of low-resolution images, we studied three mature low-resolution image recognition methods \cite{23} from the perspective of the mismatch between the spatial dimensions of high and low-resolution images. The first is to sample the high-resolution image down to the low dimensional space and match it with the low-resolution image, so as to realize the recognition of the low-resolution image. However, this method is obviously unreasonable, because the effective information contained in the high-resolution image is lost in the process of downsampling. The second method is to carry out super-resolution reconstruction of low-resolution images, and match the reconstructed images with high-resolution images in high-dimensional space to realize recognition. This method is called super-resolution image reconstruction method. But for image recognition, super-resolution reconstruction is an indirect method. Even the improved algorithm for recognition has redundant intermediate process. In addition, in order to improve the image resolution, the structure complexity of the corresponding network is generally high, the amount of parameter calculation is large, and the robustness to different resolutions is very low. In practical application, the low resolution recognition model based on super-resolution image reconstruction algorithm often needs a large number of matching high and low-resolution images, and the low-resolution images are mostly obtained by artificial subsampling, and their data distribution is inconsistent with the real environment, resulting in poor application effect of the model. The third is to map the high and low-resolution images to the same common subspace at the same time, and minimize the distance between the features of high and low resolution image  in the subspace for classification and recognition, which can more effectively improve the recognition accuracy. Ze Lu et al. \cite{24} believe that the key to using this method lies in two aspects: 1 Extract key features - they use ResNet as the backbone network to extract features; 2. find the similarity between high and low-resolution images - they use CMS to map the feature coupling on both sides to the same common feature subspace for recognition.Zhong Xin \cite{25} proposed a low-resolution vehicle recognition algorithm combining deep fusion of low-level features and multi model cooperation mechanism, which improved the recognition rate of the algorithm. In addition, shizhengyu et al. \cite{26} proposed a light discriminant self normalized neural network, which optimizes the loss function to expand the class spacing on the basis of coupling mapping to the common subspace, and introduces the scaling exponential linear element to accelerate the computational convergence, thus improving the recognition effect of low-resolution face images. The common feature subspace algorithm belongs to the direct method. Compared with the super-resolution reconstruction algorithm, there is no redundant image reconstruction process, so the structural complexity of the network is not high and the amount of parameter calculation is less. In practical application, the low-resolution recognition model based on the common feature subspace algorithm has less demand for training samples, which meets the higher requirements of hardware storage space and computing power for the deployment of automobile algorithm.
 
Therefore, we propose a two-branch residual network model based on the residual network and the common feature subspace algorithm. The feature information of high-resolution images is introduced into the training process of low-resolution networks. Due to the different processes of high and low-resolution feature extraction, we use two branch networks with different structures to map the high and low-resolution features into the same subspace. The feature extraction and classification of high and low-resolution images are independent of each other, which increases the learning freedom of network parameters. At the same time, we adopt the strategy of training high-resolution network first, and then use the idea of "distillation learning" to guide the training of low-resolution network, set the loss function as the weighted sum of the softmax cross entropy loss and the difference between the output characteristics of the two branch networks, and use the attention mechanism to make the low-resolution feature extraction network have more middle layer feature fusion modules, The low-resolution network converges to the direction of improving classification accuracy and high resolution network.
 
The current mainstream neural network model focuses on the output accuracy of the training model, and the well trained model can often provide high accuracy for a given training or test data set. Because the environmental perception technology of intelligent driving is not only realized by the machine vision perception module, but also by the joint application of multiple sensors, in a sense, the excessive pursuit of the accuracy of the network model exceeds the expected requirements for machine vision perception, but makes it put forward higher requirements for hardware storage space and computing ability, which is contrary to the original intention of the environmental perception system to pursue efficiency and real-time. Therefore, compared with other low resolution recognition methods, we pay more attention to practical application rather than blindly pursuing higher recognition accuracy. In order to realize the practical application on the intelligent vehicle platform, we use knowledge distillation to simplify the network structure, and use the low-resolution network to predict separately, which reduces the computational overhead of forward propagation.
 
According to the training data of the double branch residual network in the experimental part of this paper, the single network and the double branch network have low resolution (8 × 8) The accuracy of image recognition was 71.07\% and 72.15\%, respectively. The experimental results show that the double branch network algorithm proposed in this paper has certain advantages in recognizing low-resolution images compared with the traditional single branch network. In future research, we will further improve the application effect of the model. For example, the low-resolution image used for training can more reliably simulate the real situation; Simplify the algorithm model to meet the smaller space complexity; Further adjust the network parameters, and carry out real vehicle tests or hardware in the loop tests on the algorithm. It is hoped that our research can improve the reliability of the system in the complex and changeable road environment, and reduce the cost of hardware related to intelligent vehicles, so as to promote the application of machine vision technology in intelligent vehicles.

\newpage
\bibliographystyle{IEEEtran}
\small\bibliography{main}

\begin{IEEEbiographynophoto}{Zongcai Tan}
received the B.Eng. degree in automotive engineering from Jilin University, in 2022. He is currently working toward the MRes. degree in medical robotics and image-guided intervention at Imperial College London, London, UK. His research
interests include deep learning, computer vision, robotics and autonomous driving.
\end{IEEEbiographynophoto}

\begin{IEEEbiographynophoto}{Zhenhai Gao}
is currently a professor at Jilin University, China. Dr. Gao received his Ph.D. degree in automotive engineering from Jilin University, 2000. He was a foreign researcher of the University of Tokyo. In recent years, his research has been focused on connected \& automated vehicles, driver behavior analysis, advanced driver assistance system, and human-machine interface.

\end{IEEEbiographynophoto}
\vfill
\end{CJK}
\end{document}